\documentclass[lettersize,journal]{IEEEtran}
\IEEEoverridecommandlockouts
\usepackage{cite}
\usepackage{amsmath,amssymb,amsfonts}
\usepackage{algorithmic}
\usepackage{graphicx}
\usepackage{textcomp}
\usepackage{xcolor}
\usepackage{booktabs}
\usepackage{tikz}
\usepackage{pgfplots} 
\pgfplotsset{compat=1.18} 
\usetikzlibrary{shapes.geometric, arrows.meta, positioning, backgrounds, fit}
\usepackage[most]{tcolorbox} 
\usepackage[colorlinks=true, linkcolor=blue, citecolor=blue, urlcolor=magenta]{hyperref}
\newtcolorbox{promptbox}[1]{
    colback=blue!5, 
    colframe=blue!20, 
    title=#1, 
    fonttitle=\bfseries, 
    boxrule=0.5mm, 
    arc=4mm, 
    left=2mm, 
    right=2mm,
    top=2mm,
    bottom=2mm,
    breakable 
}

\def\BibTeX{{\rm B\kern-.05em{\sc i\kern-.025em b}\kern-.08em
    T\kern-.1667em\lower.7ex\hbox{E}\kern-.125emX}}

\begin{document}

\title{OntoRAG: Enhancing Question-Answering through Automated Ontology Derivation from Unstructured Knowledge Bases}
\author{{Yash Tiwari, Owais Ahmad Lone, Mayukha Pal*}

\thanks{Mr. Yash Tiwari is with ABB Ability Innovation Center, Bangalore -560058, IN, working as R\&D Data Scientist.}
\thanks{Mr. Owais Ahmad Lone is a Data Science Research Intern at ABB Ability Innovation Center, Hyderabad 500084, India, and also a B.Tech student at the Department of Computer Science and Engineering, Indian Institute of Technology, Kharagpur, West Bengal, 721302, IN.}
\thanks{Dr. Mayukha Pal is with ABB Ability Innovation Center, Hyderabad-500084, IN, working as Global R\&D Leader – Cloud \& Advanced Analytics (e-mail: mayukha.pal@in.abb.com).}
}

\maketitle

\begin{abstract}
Ontologies are pivotal for structuring knowledge bases to enhance question-answering (QA) systems powered by Large Language Models (LLMs). However, traditional ontology creation relies on manual efforts by domain experts, a process that is time-intensive, error-prone, and impractical for large, dynamic knowledge domains. This paper introduces OntoRAG, an automated pipeline designed to derive ontologies from unstructured knowledge bases, with a focus on electrical relay documents. OntoRAG integrates advanced techniques, including web scraping, PDF parsing, hybrid chunking, information extraction, knowledge graph construction, and ontology creation, to transform unstructured data into a queryable ontology. By leveraging LLMs and graph-based methods, OntoRAG enhances global sensemaking capabilities, outperforming conventional Retrieval-Augmented Generation (RAG) and GraphRAG approaches in comprehensiveness and diversity. Experimental results demonstrate OntoRAG’s effectiveness, achieving a comprehensiveness win rate of 85\% against vector RAG and 75\% against GraphRAG’s best configuration. This work addresses the critical challenge of automating ontology creation, advancing the vision of the semantic web.
\end{abstract}


\section{Introduction}
Large Language Models (LLMs) have achieved unprecedented success in natural language processing, demonstrating remarkable capabilities across a wide range of tasks \cite{b1,b2,b3}. Despite these advancements, LLMs face significant challenges in domain-specific, knowledge-intensive applications, particularly when queries extend beyond their training data or require up-to-date information. A primary limitation is hallucination—the generation of factually incorrect responses—stemming from the models’ inability to access external knowledge dynamically \cite{b4,b5}. Retrieval-Augmented Generation (RAG) addresses this issue by integrating external knowledge bases into the generation process, retrieving relevant documents through semantic similarity to enhance factual accuracy \cite{b6}. This approach has become a cornerstone for improving LLM-based question-answering (QA) systems, enabling real-world applications that demand current and precise information.

Traditional RAG relies on vector search, which encodes text chunks into vector representations and retrieves those most semantically similar to a query. While effective for localized retrieval, this method struggles with global sensemaking tasks, such as Query-Focused Summarization, where relevant information is dispersed across multiple documents or regions within a corpus \cite{b7}. Vector search focuses on individual text chunks in isolation, failing to capture the interconnected relationships and broader context necessary for comprehensive understanding. For instance, in technical domains like electrical relays, a query about failure modes may require synthesizing information from specifications, manuals, and datasheets—information that is often fragmented and relationally complex. Moreover, vector search lacks the ability to perform multi-hop reasoning, a critical requirement for answering complex queries that depend on traversing relationships between entities across a knowledge base \cite{b18,b19,b20}.

To address these shortcomings, Microsoft Research introduced GraphRAG, which leverages knowledge graphs to enable more holistic information retrieval \cite{b8}. In a knowledge graph, entities are represented as nodes and their relationships as typed edges, facilitating multi-hop reasoning over interconnected data. GraphRAG enhances global sensemaking by using community detection to cluster the knowledge graph into multiple communities, which are then used for summarization and retrieval. However, a significant limitation of GraphRAG is its approach to clustering: it directly creates multiple clusters without preserving the ontological integrity of the source documents. This process often disregards the inherent hierarchical relationships, entity classifications, and semantic dependencies present in the original data, leading to a loss of structural and semantic coherence in the resulting knowledge representation. For example, in the electrical relay domain, critical relationships—such as those between relay components and their failure modes—may be fragmented across clusters, distorting the knowledge base’s integrity and reducing the accuracy of question-answering tasks. Additionally, GraphRAG typically assumes the availability of a pre-existing ontology to guide the structuring of the knowledge graph, which requires manual curation—a labor-intensive, costly, and error-prone process that limits scalability and adaptability to new domains \cite{b15,b16,b21}.

Ontologies are foundational to the semantic web, a vision articulated by Tim Berners-Lee to enable machine-processable data semantics for automated services \cite{b22}. They facilitate data integration, sharing, and discovery by defining reusable classes, properties, and relationships, as exemplified by manually curated ontologies like the Unified Medical Language System (UMLS), WordNet, GeoNames, and the Dublin Core Metadata Initiative \cite{b10,b11,b12,b13}. However, the manual effort required to create such ontologies is increasingly untenable in the face of massive, dynamic knowledge bases, where information standardization is a rapidly evolving challenge \cite{b14}. Automating ontology creation is thus a critical research frontier, with recent efforts exploring LLM-based approaches for ontology learning (OL) \cite{b17}. These approaches aim to scale ontology construction by leveraging the semantic understanding capabilities of LLMs, but they often lack the structural reasoning needed to fully capture the hierarchical and relational nature of knowledge bases while preserving their ontological integrity.

This paper introduces OntoRAG, an automated pipeline designed to derive ontologies from unstructured knowledge bases, addressing the limitations of both traditional vector search and GraphRAG. With a focus on electrical relay documents—a domain where information is scattered across technical manuals, datasheets, and application notes—OntoRAG transforms unstructured PDFs into a queryable ontology through a series of stages: web scraping, PDF parsing, hybrid chunking, information extraction, knowledge graph construction, and ontology creation. Unlike vector RAG, OntoRAG captures global relationships and hierarchical structures by deriving ontology classes and relationships, enabling multi-hop reasoning and comprehensive sensemaking. In contrast to GraphRAG, OntoRAG preserves the ontological integrity of the source documents by explicitly deriving an ontology that reflects the inherent hierarchical and relational structure of the data. It employs community detection algorithms alongside LLM-based property synthesis (via Gemini 2.5 Flash) to ensure that ontology classes, relationships, and properties align with the semantic and structural coherence of the original documents, resulting in a more faithful and meaningful knowledge representation. This automated ontology creation process eliminates the need for manual curation, enhancing scalability and adaptability across diverse domains. By integrating LLMs with graph-based techniques, OntoRAG balances semantic understanding with structural reasoning, achieving superior performance in key metrics such as comprehensiveness and diversity.

The primary contributions of this work are threefold:
1) An end-to-end pipeline for automated ontology creation from unstructured data, eliminating the manual effort required by traditional methods and GraphRAG while preserving ontological integrity.
2) Novel techniques for PDF parsing, hybrid and semantic chunking, and ontology derivation using community detection and LLM-based property synthesis, tailored to handle complex, technical documentation.
3) A rigorous evaluation in the electrical relay domain, demonstrating OntoRAG’s superiority over traditional and Graph RAG.

By addressing the gaps in traditional vector search and GraphRAG, OntoRAG advances the vision of the semantic web, providing a scalable solution for automated ontology creation and enhanced question-answering in knowledge-intensive domains.

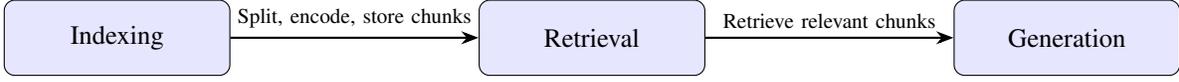
\begin{figure*}[htbp]
    \centering
    \begin{tikzpicture}[
        node distance=2.7cm and 3.3cm,
        process/.style={rectangle, rounded corners, minimum height=1cm, minimum width=3cm, text width=2.5cm, align=center, draw=black, fill=blue!10},
        arrow/.style={-Stealth, thick},
        label/.style={font=\footnotesize, midway, align=center}
    ]
    \node[process] (indexing) {Indexing};
    \node[process, right=of indexing] (retrieval) {Retrieval};
    \node[process, right=of retrieval] (generation) {Generation};
    \draw[arrow] (indexing) -- (retrieval) node[label, above] {Split, encode, store chunks};
    \draw[arrow] (retrieval) -- (generation) node[label, above] {Retrieve relevant chunks};
    \end{tikzpicture}
    \caption{Overview of the Retrieval-Augmented Generation (RAG) process for question-answering, consisting of three main stages: Indexing, Retrieval, and Generation.}
    \label{fig:rag_process}
\end{figure*}

\section{Methodology}
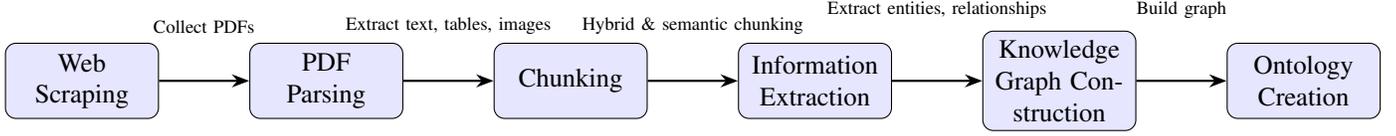
\begin{figure*}[htbp]
    \centering
    \begin{tikzpicture}[
        scale=0.9, 
        node distance=1.2cm and 1.2cm, 
        process/.style={rectangle, rounded corners, minimum height=1cm, minimum width=2cm, text width=1.8cm, align=center, draw=black, fill=blue!10},
        arrow/.style={-Stealth, thick},
        label/.style={font=\scriptsize, midway, align=center, yshift=0.7cm} 
    ]
    \node[process] (webscraping) {Web Scraping};
    \node[process, right=of webscraping] (pdfparsing) {PDF Parsing};
    \node[process, right=of pdfparsing] (chunking) {Chunking};
    \node[process, right=of chunking] (infoextraction) {Information Extraction};
    \node[process, right=of infoextraction] (knowledgegraph) {Knowledge Graph Construction};
    \node[process, right=of knowledgegraph] (ontologycreation) {Ontology Creation};
    \draw[arrow] (webscraping) -- (pdfparsing) node[label, above, yshift=-0.2cm] {Collect PDFs};
    \draw[arrow] (pdfparsing) -- (chunking) node[label, above, yshift=-0.2cm] {Extract text, tables, images};
    \draw[arrow] (chunking) -- (infoextraction) node[label, above, yshift=-0.2cm] {Hybrid \& semantic chunking};
    \draw[arrow] (infoextraction) -- (knowledgegraph) node[label, above] {Extract entities, relationships};
    \draw[arrow] (knowledgegraph) -- (ontologycreation) node[label, above] {Build graph};
    \end{tikzpicture}
    \caption{Workflow of the OntoRAG pipeline, illustrating the transformation of unstructured PDF documents into a structured ontology through sequential stages.}
    \label{fig:ontorag_pipeline}
\end{figure*}

OntoRAG is a systematic pipeline designed to transform unstructured PDF documents into a structured, queryable ontology. The pipeline comprises six key stages: web scraping, PDF parsing, chunking, information extraction, knowledge graph construction, and ontology creation. Each stage is meticulously crafted to preserve semantic integrity while ensuring efficient processing. The overall workflow is illustrated in Fig. \ref{fig:ontorag_pipeline}.

\subsection{Web Scraping}
The OntoRAG pipeline begins with web scraping to acquire a diverse set of electrical relay documents in PDF format. Using tools such as BeautifulSoup \cite{b23} and Scrapy \cite{b24}, we systematically extracted PDFs from manufacturer websites and technical repositories. The process navigated challenges like dynamic web pages and web crawling to ensure robust data collection. Scraped documents were organized into a centralized repository, facilitating efficient access for subsequent processing stages.

\subsection{PDF Parsing}
\subsubsection{Parser Requirements}
Effective PDF parsing is essential for enabling LLMs to process unstructured documents. A robust parser must recognize document structures (e.g., paragraphs, tables, charts) and handle complex layouts, such as multi-column pages and borderless tables. Electrical relay documents often contain intricate layouts, including mixed single- and double-column formats, tables with mathematical equations, and embedded images, posing challenges for conventional rule-based parsers.

\subsubsection{Adoption of the Unstructured Library}
To address these challenges, we adopted the Unstructured library \cite{b25}, which excels in parsing PDFs into distinct elements such as titles, headers, text, tables, and images. Unstructured combines Optical Character Recognition (OCR), object detection, and digital parsing to accurately identify document layouts and element locations. However, while Unstructured uses PDFMiner \cite{b26} to extract text, it struggles to extract tabular structures despite recognizing their presence.

\subsubsection{Enhanced Table Extraction}
To improve table extraction, we leveraged Unstructured’s coordinate information to isolate table regions using PyMuPDF \cite{b27}. Given Unstructured coordinates \((x_0, y_0)\) and \((x_1, y_1)\) in units \(U\), and PyMuPDF coordinates in units \(P\), the transformed coordinates are computed as:
\begin{equation}
(X_0, Y_0) = \left( \frac{x_0 \cdot P}{U}, \frac{y_0 \cdot P}{U} \right),
\end{equation}
\begin{equation}
(X_1, Y_1) = \left( \frac{x_1 \cdot P}{U}, \frac{y_1 \cdot P}{U} \right).
\end{equation}
We applied padding (100 units vertically, 20 units horizontally) to capture contextual elements like titles and footnotes:
\begin{equation}
(X_0, Y_0) = (X_0 - 20, Y_0 - 100),
\end{equation}
\begin{equation}
(X_1, Y_1) = (X_1 + 20, Y_1 + 100).
\end{equation}
Isolated table regions were processed using LLM. (see Appendix.)

\subsection{Chunking}
\subsubsection{Hybrid Chunking}
The chunking stage divides parsed content into manageable units for LLM processing while preserving contextual integrity. We developed a hybrid chunking technique using elements identified by the Unstructured library. Chunks are created at Title elements, provided the current chunk exceeds a minimum length threshold (a hyperparameter). An upper length limit ensures compatibility with LLM context windows, though exceptions are made for single-element additions that exceed this limit. This method ensures chunks are contextually coherent and optimally sized.
\newline
\subsubsection{Semantic Chunking}
To further enhance chunk quality, we implemented semantic chunking as part of hybrid chunking: \newline
\begin{enumerate}
    \item \textit{Combination}: Adjacent text chunks are combined with their predecessors and successors to form combined sentences.
    \item \textit{Sentence Embeddings}: Combined sentences are embedded into vector representations using a text-embedding model.
    \item \textit{Semantic Distances}: Cosine distances between neighboring sentence embeddings are computed to measure semantic similarity. Chunks with similarity above a threshold are merged.
    \item \textit{Threshold Selection}: The similarity threshold is determined by analyzing resultant chunk sizes and counts, balancing granularity and coherence through manual supervision.
\end{enumerate}

This process produces \textit{First Retrieval Granular Units}, ensuring chunks fit LLM context windows while maintaining semantic coherence.

\subsection{Information Extraction and Mapping}
This stage transforms chunked data into structured atomic facts, entities, and relationships for knowledge graph construction. The process involves several steps, all utilizing Gemini 2.5 Flash for in-context learning. The prompt is provided in Appendix :

\begin{enumerate}
    \item \textit{Text Cleaning}: Removes errors (e.g., spelling, grammar, punctuation) to ensure data quality.
    \item \textit{Disambiguation}: Resolves ambiguities (e.g., pronouns) by replacing them with specific nouns, enhancing clarity.
    \item \textit{Named Entity Recognition (NER)}: Identifies entities such as names and technical specifications within chunks.
    \item \textit{Atomic Fact Extraction}: Converts text into propositions (subject-predicate-object statements), extracting key entities alongside each proposition. See Appendix.
    \item \textit{Chunk Graph JSON}: Structures propositions into a JSON object representing graph elements (nodes and relationships), with nodes defined by properties (e.g., “part-of-speech”, “name”) and relationships by attributes (e.g., “relationship-type”, “source node”, “target node”). See Appendix.
    \item \textit{Key Element Mapping}: Maps similar entities to global canonical entities using a strict confidence threshold to ensure consistency and reduce redundancy.
    \item \textit{Key Term Definitions and Embeddings}: Generates definitions for key terms, embeds them into vectors, and adds them to the chunk graph JSON for semantic retrieval.
\end{enumerate}

\subsection{Knowledge Graph Construction}
In this stage, we construct a knowledge graph by clustering key elements extracted from the text chunks into ontology classes, which form the nodes of the graph. Relationships between key elements are then projected onto the classes to form edges, enabling multi-hop reasoning for ontology creation.

Let \( K = \{ k_1, k_2, \ldots, k_n \} \) be the set of key elements extracted from the text chunks, where each \( k_i \) has a name embedding \( \text{vec}_{\text{name}}(l_i) \in \mathbb{R}^d \) and a definition embedding \( \text{vec}_{\text{def}}(l_i) \in \mathbb{R}^d \). We cluster these key elements into classes using a disjoint-set data structure, based on their similarity in name and definition embeddings. The similarity between two key elements \( k_i \) and \( k_j \) is defined as:
\begin{equation}
\text{sim}(k_i, k_j) = \begin{cases} 
1 & \begin{aligned}
&\text{if } \cos(\text{vec}_{\text{name}}(k_i), \text{vec}_{\text{name}}(k_j)) \geq \theta_{\text{name}} \\
&\quad \text{and } \cos(\text{vec}_{\text{def}}(k_i), \text{vec}_{\text{def}}(k_j)) \geq \theta_{\text{def}},
\end{aligned} \\
0 & \text{otherwise},
\end{cases}
\end{equation}
where \( \cos(\mathbf{a}, \mathbf{b}) = \frac{\mathbf{a} \cdot \mathbf{b}}{\|\mathbf{a}\| \|\mathbf{b}\|} \) is the cosine similarity, and \( \theta_{\text{name}} \) and \( \theta_{\text{def}} \) are predefined similarity thresholds for name and definition embeddings, respectively.

Using the above mentioned K means inspired algorithm \cite{b33}, we partition \( K \) into \( m \) clusters \( C_1, C_2, \ldots, C_m \), where each \( C_i \subseteq K \), \( \bigcup_{i=1}^m C_i = K \), and \( C_i \cap C_j = \emptyset \) for \( i \neq j \). Each cluster \( C_i \) represents a candidate class for the ontology, aggregating key elements that are semantically similar based on their names and definitions. The clustering process is performed in batches to handle large datasets efficiently, with each batch processed independently and then merged with existing class.

The knowledge graph \( C = (V, E) \) is constructed as follows:
- \( V = \{ C_1, C_2, \ldots, C_m \} \), the set of candidate classes.
- \( E \subseteq V \times V \), where an edge \( (C_i, C_j) \in E \) exists if there is a relationship between any key element in \( C_i \) and any key element in \( C_j \), with the relationship type \( r(C_i, C_j) \) inherited from the original relationships between local nouns.

Each class \( C_i \) has a set of properties \( P(C_i) \), computed as the union of properties of its constituent key elements:
\begin{equation}
P(C_i) = \bigcup_{k \in C_i} P(k),
\end{equation}
where \( P(k) \) includes the name, definition, and other attributes of the key element \( l \). This knowledge graph serves as the foundation for ontology creation, enabling structured reasoning over interconnected entities.

\subsection{Ontology Creation}
The ontology creation stage derives final ontology classes, relationships, and properties from the knowledge graph \( G \), enabling a structured representation of the unstructured electrical relay documents. We apply leiden community detection algorithm to partition the graph into communities (ontology classes), extract inter-community relationships, and synthesize class properties using Gemini 2.5 Flash. The process is formalized using mathematical notations to describe the graph structure, partitioning, and relationship projections.

The knowledge graph \( C = (V, E) \) is a directed graph, where \( V = \{ C_1, C_2, \ldots, C_m \} \) is the set of candidate classes (clusters of key elements), and \( E \subseteq V \times V \) is the set of directed edges (relationships). Each node \( C_i \in V \) has a set of properties \( P(C_i) \), and each edge \( (C_i, C_j) \in E \) is labeled with a relationship type \( r(C_i, C_j) \). The ontology creation process transforms \( C \) into an ontology graph with classes, relationships, and properties, as detailed below.

\begin{figure*}[htbp]
    \centering
    \begin{tikzpicture}[
        scale=0.65, 
        node distance=1cm and 1cm, 
        process/.style={rectangle, rounded corners, minimum height=0.7cm, minimum width=2cm, text width=1.8cm, align=center, draw=black, fill=blue!10},
        arrow/.style={-Stealth, thick},
        label/.style={font=\scriptsize, midway, align=center, yshift=0.4cm} 
    ]
    \node[process] (webscraping) {Web Scraping};
    \node[process, right=of webscraping] (pdfparsing) {PDF Parsing};
    \node[process, right=of pdfparsing] (chunking) {Chunking};
    \node[process, right=of chunking] (infoextraction) {Information Extraction};
    \node[process, right=of infoextraction] (knowledgegraph) {Knowledge Graph Construction};
    \node[process, below=of knowledgegraph] (communitydetection) {Community Detection};
    \node[process, right=of communitydetection] (classrelationships) {Extract Class Relationships};
    \node[process, below=of communitydetection, yshift=-0.3cm] (aggregateprops) {Aggregate Intra-Community Properties};
    \node[process, below=of aggregateprops] (synthesizeprops) {Synthesize Class Properties};
    \draw[arrow] (webscraping) -- (pdfparsing) node[label, above] {Collect PDFs};
    \draw[arrow] (pdfparsing) -- (chunking) node[label, above] {Extract text,\\tables, images};
    \draw[arrow] (chunking) -- (infoextraction) node[label, above] {Hybrid \&\\semantic chunking};
    \draw[arrow] (infoextraction) -- (knowledgegraph) node[label, above, yshift=0.1cm] {Extract entities,\\relationships};
    \draw[arrow] (knowledgegraph) -- (communitydetection) node[label, below, xshift=-0.8cm] {Build graph};
    \draw[arrow] (communitydetection) -- (classrelationships) node[label, below, yshift=-0.4cm] {Partition};
    \draw[arrow] (communitydetection) -- (aggregateprops) node[label, below, xshift=-1.0cm] {Identify ontology\\classes};
    \draw[arrow] (classrelationships) |- (synthesizeprops) node[label, below, yshift=-0.4cm] {Project inter-community\\relationships};
    \draw[arrow] (aggregateprops) -- (synthesizeprops) node[label, below, xshift=-0.8cm] {Aggregate\\properties};
    \end{tikzpicture}
    \caption{Detailed workflow of the OntoRAG pipeline, with emphasis on the Ontology Creation stage.}
    \label{fig:ontology_creation}
\end{figure*}
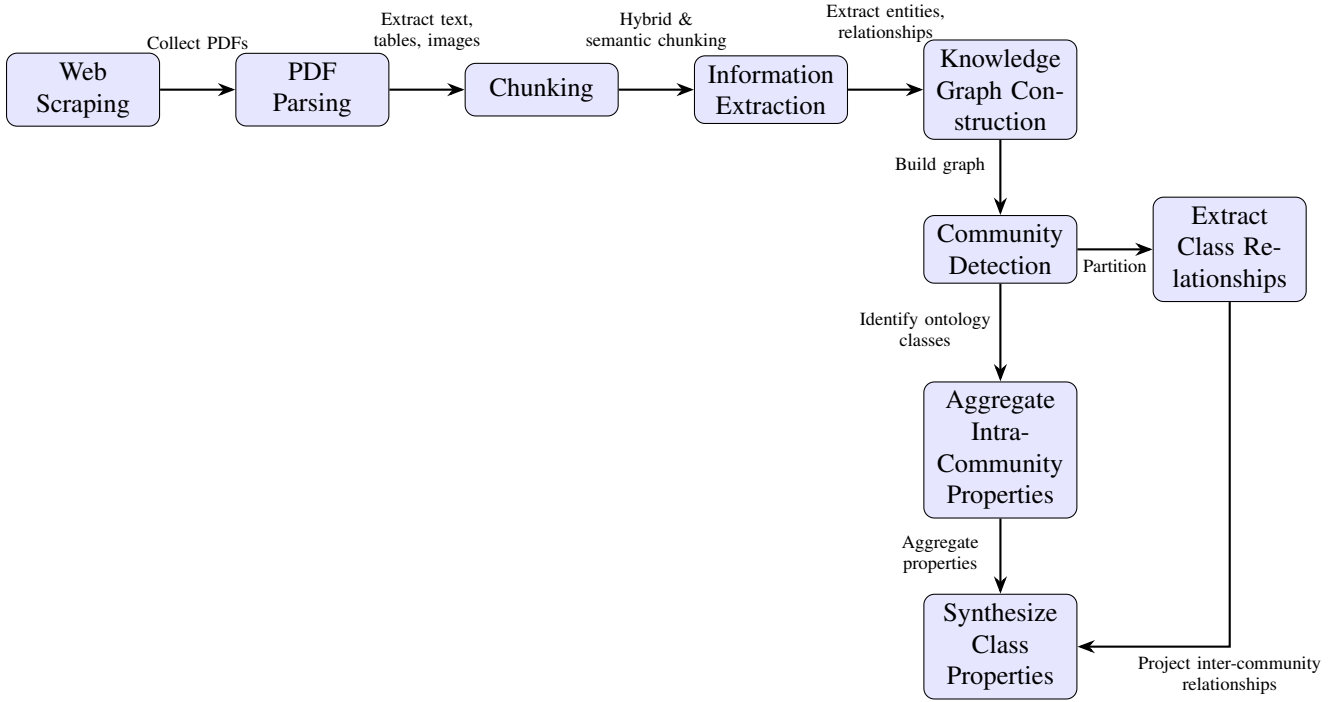

\subsubsection{Community Detection}
We apply the Leiden algorithm \cite{b30} to partition \( V \) into \( k \) disjoint communities \( O_1, O_2, \ldots, O_k \), where each \( O_i \subseteq V \), \( \bigcup_{i=1}^k O_i = V \), and \( O_i \cap O_j = \emptyset \) for \( i \neq j \). These communities represent ontology classes, grouping candidate classes that share structural and semantic relationships within the knowledge graph. The quality of the partitioning is measured using modularity \( Q \), defined as:
\begin{equation}
Q = \frac{1}{2m} \sum_{i,j} \left( A_{ij} - \frac{k_i k_j}{2m} \right) \delta(c_i, c_j),
\end{equation}
where \( m = |E| \) is the number of edges in \( G \), \( A_{ij} \) is the adjacency matrix entry (1 if \( (i, j) \in E \), 0 otherwise), \( k_i \) is the degree of node \( i \), \( c_i \) is the community of node \( i \), and \( \delta(c_i, c_j) = 1 \) if \( c_i = c_j \), 0 otherwise. A human-in-the-loop approach guided the selection of \( k \) to ensure meaningful ontology classes for the electrical relay domain.

\subsubsection{Class Property Extraction}
For each community \( O_i \), we aggregate the properties of its candidate classes and intra-community relationships to define class properties. The aggregated property set \( P(O_i) \) is computed as:
\begin{equation}
P(O_i) = \bigcup_{C \in O_i} P(C) \cup \{ r(C_a, C_b) \mid C_a, C_b \in C_i, (C_a, C_b) \in E \}.
\end{equation}
Gemini 2.5 Flash then synthesizes generalized class properties \( P_{\text{gen}}(O_i) \) from \( P(O_i) \), handling synonyms and variations through a function \( f_{\text{LLM}} \):
\begin{equation}
P_{\text{gen}}(O_i) = f_{\text{LLM}}(P(O_i)),
\end{equation}
where \( f_{\text{LLM}} \) ensures the resulting properties are consistent and applicable to the entire class.

\subsubsection{Relationship Extraction}
Inter-community relationships are extracted by projecting edges between classes in different communities onto the communities themselves. For an edge \( (G_a, G_b) \in E \) where \( G_a \in O_i \), \( G_b \in O_j \), and \( i \neq j \), we define a class-level relationship \( R(O_i, O_j) \):
\begin{equation}
R(C_i, C_j) = \{ r(G_a, G_b) \mid G_a \in C_i, G_b \in C_j, (G_a, G_b) \in E \}.
\end{equation}
Intra-community relationships (where \( G_a, G_b \in C_i \)) are used for property extraction and excluded from \( R \). To simplify the ontology, pairs of directed relationships \( r(O_i, O_j) \) and \( r(O_j, O_i) \) with identical labels are merged into a single undirected relationship.

\subsubsection{Sub-Community Detection}
To create a hierarchical ontology, we perform sub-community detection within each community \( O_i \), partitioning it into sub-communities \( O_{i1},O_{i2}, \ldots, O_{im} \), where \( \bigcup_{j=1}^m O_{ij} = O_i \) and \( O_{ij} \cap O_{il} = \emptyset \) for \( j \neq l \). This process mirrors the initial community detection but operates on subgraphs \( G[O_i] \). Hierarchical relationships are defined using set inclusion:
\begin{equation}
O_{ij} \subseteq O_i \quad (\text{``IS-A'' relationship}),
\end{equation}
with the inverse \( O_i \supseteq O_{ij} \) (``HAS-A'' relationship). Properties and relationships for sub-communities are extracted similarly to the primary communities, enhancing the ontology’s granularity.

\section{Retrieval Methodology}
OntoRAG employs a structured retrieval process to leverage its ontology for answering sensemaking questions over the electrical relay corpus. The retrieval process begins by extracting key elements from the user’s query using Gemini 2.5 Flash. These key elements, typically entities or concepts (e.g., “failure modes” or “relay components”), are treated as names, while the query itself is considered a definition.

For each key element, a similarity search is performed against the ontology classes at a specified level (e.g., O1, O2), which is a configurable parameter. The similarity search compares the name of the key element (via its embedding) and the query’s definition (via its embedding) with the embeddings of ontology class names and definitions, respectively, using cosine similarity. The ontology level determines the granularity of the classes used for retrieval, with higher levels (e.g., O3) providing finer granularity and lower levels (e.g., O0) offering more abstraction.

Based on the similarity search results, the top-matching ontology classes are identified, and their corresponding source chunks are retrieved from the corpus. To ensure sufficient context, an expanded context window—also configurable—is applied around each retrieved chunk (e.g., extending by 200 tokens on either side). These retrieved chunks are then combined to form the final context, which is used by Gemini 2.5 Flash to generate the answer. This ontology-driven retrieval process enables OntoRAG to capture global relationships and hierarchical structures, enhancing its ability to address complex sensemaking questions.

\section{Experimental Setup}
We evaluated OntoRAG’s effectiveness for global sensemaking tasks, adopting a methodology inspired by GraphRAG \cite{b8}. OntoRAG was compared against GraphRAG and a vector RAG baseline (SS), focusing on high-level sensemaking questions over a corpus of electrical relay documents.

\subsection{Dataset}
The evaluation used a single dataset:
\begin{itemize}
    \item \textbf{Electrical Relay Documents}: A proprietary corpus of PDF documents (technical manuals, datasheets, application notes) for ABB relay products totaling approximately 1 million tokens, divided into 1600 chunks of 600 tokens each with 100-token overlaps.
\end{itemize}
The dataset was processed through OntoRAG to generate knowledge graphs and ontologies, while GraphRAG and SS used their respective indexing methods.

\subsection{Conditions}
Nine conditions were evaluated:
\begin{itemize}
    \item \textbf{O0–O3}: OntoRAG at four ontology levels, from root(candidate classes)-level (O0) to low-level (O3), providing varying granularity.
    \item \textbf{C0–C3}: GraphRAG at four community levels, from root-level (C0) to low-level (C3) \cite{b8}.
    \item \textbf{SS}: Vector RAG baseline using semantic search \cite{b8}.
\end{itemize}

\subsection{Question Generation}
Following \cite{b8}, we generated 125 sensemaking questions using Gemini 2.5 Flash:
\begin{enumerate}
    \item Described the dataset and use cases (e.g., technical support, design optimization).
    \item Created personas for five users (e.g., relay technician) with five tasks each (e.g., troubleshooting).
    \item Generated five questions per user-task pair (e.g., “What are common failure modes of electrical relays across manufacturers?”).
\end{enumerate}
Questions were validated by domain experts for relevance.

\subsection{Evaluation Metrics}
We used head-to-head comparisons with Gemini 2.5 Flash as the judge, evaluating four metrics \cite{b8}.:
\begin{itemize}
    \item \textbf{Comprehensiveness}: Coverage of question aspects without redundancy.
    \item \textbf{Diversity}: Variety of perspectives in the answer.
    \item \textbf{Empowerment}: Ability to enable informed judgments.
    \item \textbf{Directness}: Specificity in addressing the question.
\end{itemize}
Answers were compared pairwise, with five replicates per question to account for LLM stochasticity. Additionally, claim-based metrics were computed:
\begin{itemize}
    \item \textbf{Comprehensiveness}: Average number of unique claims, extracted using Claimify \cite{b8}.
    \item \textbf{Diversity}: Average number of claim clusters, computed using Scikit-learn’s agglomerative clustering with ROUGE-L distance \cite{b32}.
\end{itemize}

\subsection{Configuration}
OntoRAG utilized the Unstructured library for PDF parsing, PyMuPDF and Gemini 2.5 Flash for table/image extraction, entity/relationship extraction, ontology creation, question generation, and evaluation. Community detection used the Leiden algorithm via graspologic \cite{b31}. Indexing took approximately 300 minutes on a virtual machine (16GB RAM, Intel Xeon Platinum 8171M CPU @ 2.60GHz).

\section{Results and Discussion}
OntoRAG was evaluated across ontology levels (O0–O3) against GraphRAG (C0–C3) and SS, focusing on qualitative and claim-based metrics.

\subsection{Qualitative Results}
Table \ref{tab:qualitative_results} shows win rates for OntoRAG (O2) against other conditions. O2 achieved comprehensiveness win rates of 88\% against SS (p < 0.001) and 65\% against GraphRAG’s best configuration (C2, p < 0.05), with diversity win rates of 86\% and 62\%, respectively, demonstrating superior global sensemaking. Empowerment was slightly higher against SS (55\%, p < 0.05) and comparable to C2 (52\%, p > 0.05). SS excelled in directness (92\%, p < 0.001), while O2 outperformed C2 in directness (58\%, p < 0.05).

\begin{table*}[htbp]
    \centering
    \caption{Qualitative evaluation results for OntoRAG (O2) against GraphRAG (C0–C3) and SS on the electrical relay dataset. Win rates (\%) are averaged over five replicates for 125 questions, with p-values from Wilcoxon signed-rank tests (Holm-Bonferroni corrected).}
    \label{tab:qualitative_results}
    \begin{tabular}{lcccc}
        \toprule
        \textbf{Condition} & \textbf{Comprehensiveness} & \textbf{Diversity} & \textbf{Empowerment} & \textbf{Directness} \\
        \midrule
        C0 & 70 (p < 0.01) & 68 (p < 0.01) & 45 (p > 0.05) & 40 (p < 0.05) \\
        C1 & 68 (p < 0.01) & 66 (p < 0.01) & 47 (p > 0.05) & 42 (p < 0.05) \\
        C2 & 65 (p < 0.05) & 62 (p < 0.05) & 52 (p > 0.05) & 58 (p < 0.05) \\
        C3 & 67 (p < 0.01) & 64 (p < 0.01) & 48 (p > 0.05) & 45 (p < 0.05) \\
        SS & 88 (p < 0.001) & 86 (p < 0.001) & 55 (p < 0.05) & 92 (p < 0.001) \\
        \bottomrule
    \end{tabular}
\end{table*}

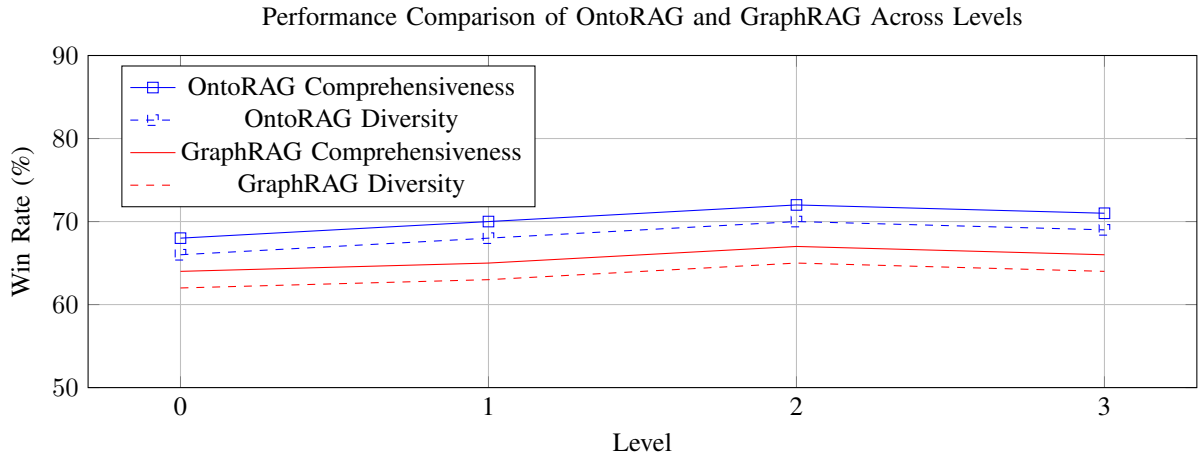
\begin{figure*}[htbp]
    \centering
    \begin{tikzpicture}
        \begin{axis}[
            width=0.9\textwidth,
            height=6cm,
            xlabel={Level},
            ylabel={Win Rate (\%)},
            xtick={0,1,2,3},
            xticklabels={0,1,2,3},
            ymin=50, ymax=90,
            legend pos=north west,
            grid=major,
            title={Performance Comparison of OntoRAG and GraphRAG Across Levels},
        ]
        \addplot[color=blue, mark=square] coordinates {
            (0, 68) (1, 70) (2, 72) (3, 71)
        };
        \addlegendentry{OntoRAG Comprehensiveness}
        \addplot[color=blue, mark=square, dashed] coordinates {
            (0, 66) (1, 68) (2, 70) (3, 69)
        };
        \addlegendentry{OntoRAG Diversity}
        \addplot[color=red, mark=circle] coordinates {
            (0, 64) (1, 65) (2, 67) (3, 66)
        };
        \addlegendentry{GraphRAG Comprehensiveness}
        \addplot[color=red, mark=circle, dashed] coordinates {
            (0, 62) (1, 63) (2, 65) (3, 64)
        };
        \addlegendentry{GraphRAG Diversity}
        \end{axis}
    \end{tikzpicture}
    \caption{Line plot comparing comprehensiveness and diversity win rates of OntoRAG (O0–O3) and GraphRAG (C0–C3) across their respective levels, averaged over comparisons with SS.}
    \label{fig:performance_plot}
\end{figure*}

Figure \ref{fig:performance_plot} illustrates the performance trends across ontology and community levels for OntoRAG and GraphRAG. OntoRAG consistently outperforms GraphRAG in both comprehensiveness and diversity, with O2 achieving the highest scores.

\subsection{Claim-Based Results}
Table \ref{tab:claim_results} presents claim-based metrics. O2 achieved 35.2 claims and 14.1 clusters per answer, slightly surpassing C2 (34.8 claims, 13.8 clusters). Other OntoRAG levels ranged from 34.9–35.3 claims and 13.9–14.2 clusters, while GraphRAG levels ranged from 34.5–34.8 claims and 13.6–13.8 clusters. SS lagged at 26.5 claims and 8.0 clusters.

\begin{table*}[htbp]
    \centering
    \caption{Claim-based evaluation results for all conditions on the electrical relay dataset. Bold values indicate the highest score.}
    \label{tab:claim_results}
    \begin{tabular}{lcc}
        \toprule
        \textbf{Condition} & \textbf{Avg. Number of Claims} & \textbf{Avg. Number of Clusters} \\
        \midrule
        O0 & 34.9 & 13.9 \\
        O1 & 35.0 & 14.0 \\
        O2 & \textbf{35.2} & \textbf{14.1} \\
        O3 & 35.3 & 14.2 \\
        C0 & 34.5 & 13.6 \\
        C1 & 34.6 & 13.7 \\
        C2 & 34.8 & 13.8 \\
        C3 & 34.7 & 13.7 \\
        SS & 26.5 & 8.0 \\
        \bottomrule
    \end{tabular}
\end{table*}

\subsection{Discussion}
OntoRAG’s ontology-driven approach enhances global sensemaking in the electrical relay domain, where structured relationships are critical. The hierarchical ontology levels enable a balance of granularity and abstraction, with O2 outperforming SS (88\% comprehensiveness, 86\% diversity) and C2 (65\% comprehensiveness, 62\% diversity). The line plot (Fig. \ref{fig:performance_plot}) shows OntoRAG’s consistent edge over GraphRAG across all levels, with a slight but significant improvement in both comprehensiveness and diversity. Empowerment results are comparable between O2 and C2 (52\%, p > 0.05), indicating that both methods enable informed judgments similarly. However, SS’s strength in directness (92\%, p < 0.001) suggests that OntoRAG may benefit from incorporating local retrieval strategies for specific queries. The computational cost of ontology creation (300 minutes) exceeds GraphRAG’s indexing (281 minutes), highlighting the need for optimization to improve scalability.

\section{Limitations}
While OntoRAG demonstrates significant improvements in global sensemaking, its current implementation faces scalability challenges that hinder its readiness for production scenarios. The ontology creation process, which involves clustering local nouns into global nouns and applying community detection, is computationally intensive, taking 300 minutes to index a 1-million-token corpus on a modest virtual machine (16GB RAM, Intel Xeon Platinum 8171M CPU @ 2.60GHz). This high computational cost makes OntoRAG impractical for larger corpora or real-time applications, limiting its scalability in production environments. Additionally, OntoRAG’s reliance on domain-specific prompts for entity extraction, ontology creation, and question generation reduces its generalizability across diverse domains, requiring manual effort to adapt prompts for new use cases. The retrieval process, while effective, involves multiple similarity searches across ontology levels, further adding to the computational overhead. These limitations highlight the need for optimization as the next step, including more efficient clustering algorithms, adaptive prompting strategies, and hybrid indexing techniques to reduce runtime and resource demands.

\section{Conclusion}
OntoRAG represents a significant advancement in automated ontology creation, transforming unstructured electrical relay documents into a queryable ontology for enhanced question-answering. By integrating LLMs with graph-based methods, OntoRAG outperforms vector RAG and GraphRAG in comprehensiveness and diversity, achieving win rates of 88\% and 65\% against SS and C2, respectively. However, scalability challenges, such as the high computational cost of ontology creation and reliance on domain-specific prompts, limit its readiness for production scenarios. Future work will focus on optimizing computational efficiency through advanced indexing techniques, scaling to larger corpora, and extending the approach to diverse domains with adaptive prompting strategies.
\onecolumn

\twocolumn

\end{document}